\pdfoutput=1
\documentclass[a4paper]{article}    
\usepackage[numbers]{natbib}
\setcitestyle{square,aysep={},yysep={;}} 
\usepackage{times}
\usepackage{helvet}
\usepackage{url}
\usepackage{graphicx}

\usepackage{booktabs} 
\usepackage{enumerate}
\usepackage{verbatim}

\usepackage{algorithm,algorithmic,  xcolor}
\usepackage[fleqn]{amsmath}
\usepackage{times, epsfig, theorem, amssymb,graphicx, wrapfig}

\newlength{\leftmargindeduction}
\usepackage{lineno}  
\begin{document}

 \title{A transformer boosted UNet for smoke segmentation  in complex backgrounds in multispectral LandSat imagery}
\author{Jixue Liu,  Jiuyong Li,  Stefan Peters, Liang Zhao \\
UniSA STEM, University of South Australia \\
\{jixue.liu, jiuyong.li, stefan.peters, liang.zhao\}@unisa.edu.au}

\maketitle

 \begin{abstract}
Many studies have been done to detect smokes from satellite imagery. However, these prior methods are not still effective in detecting various smokes in complex backgrounds. Smokes present challenges in detection due to variations in density, color, lighting, and backgrounds such as clouds, haze, and/or mist, as well as the contextual nature of thin smoke. This paper addresses these challenges by proposing a new segmentation model called  VTrUNet which consists of a virtual band construction module to capture spectral patterns and a transformer boosted UNet to capture long range contextual features. The model takes imagery of six bands: red, green, blue, near infrared, and two shortwave infrared bands as input. To show the advantages of the proposed model, the paper presents extensive results for various possible model architectures improving UNet and draws interesting conclusions including that adding more modules to a model does not always lead to a better performance. The paper also compares the proposed model with very recently proposed and related models for smoke segmentation and shows that the proposed model performs the best and makes significant improvements on prediction performances. 
\end{abstract}


\section{Introduction}

Wildfires have caused significant losses in environment, economy and people's lives. Direct detection of wild fires is challenging when they are small or fire locations are remote. Fire detection via smoke detection is an effective method because smokes spread and rise faster and are important signals of wildfires. Smoke detection from satellite imagery has the advantages of covering remote areas and being daylight independent and it has attracted a lot of research.  

\begin{figure*}[t]
\begin{center}
(a)\hspace{2.5cm}(b)\hspace{2.5cm}(c)\hspace{2.5cm}(d)\\
\includegraphics[scale=0.50]{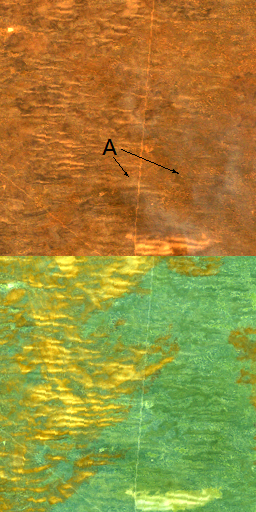} 
\includegraphics[scale=0.50]{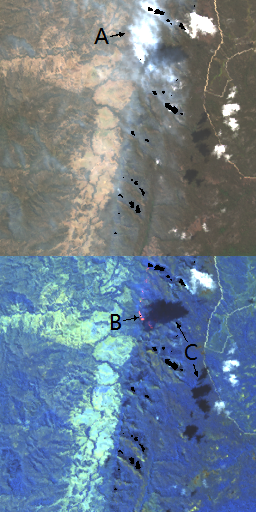}
\includegraphics[scale=0.50]{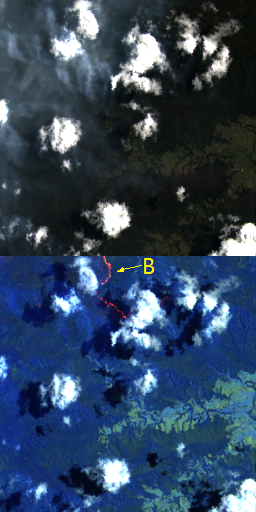} 
\includegraphics[scale=0.50]{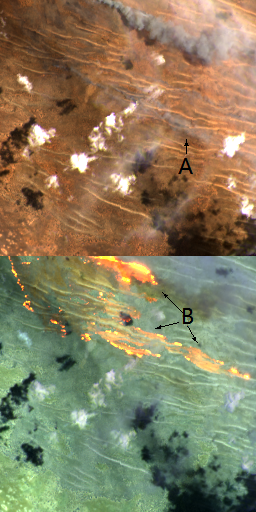} 
\caption{\label{fig:relativeness} 
Complexities of smokes in satellite images. In each part, the top is RGB, and the bottom is the false color image for bands of NIR, SWIR1, SWIR2.
(a) shows thin smoke A over clear land. (b) shows thick smoke A  surrounded by thin smoke/haze and accompanied by active fire B, cloud and cloud shadows  C in the bottom. (c) land cover is unclear at all in the RGB image, and fire front line and cloud shadows on the false color image. (d) wide and narrow black smokes A along the fire front line B and cloud.  
}
\end{center}
\end{figure*}

The work on smoke detection varies from satellite imagery \citep{ClassificationTree-smoke-2020, smoke-flumes-2021a, zhaoliang22, VIIRs-Landsat-imgs2023, BoucaNet2023},  to UAV (unmanned aerial vehicle) imagery \cite{UAV-YOLOv7-2023}, land camera imagery \citep{SelfAttentionStrawSmoke2022,AttentionCNNsmoke-fog2021, cross-contrast-swin-transformer-2023} (see \citep{survey-smoke-detect-tech2022} for more)  and videos \citep{video-smoke-recog-atten-Tao2024CEA,multimodal-smoke-landcam-2023,fire-recog-deepCNN-2024,Video-DeformableConv-SmokeRecog-Tao2024ESA}. Our work is on smoke detection from satellite imagery.
 
Satellite imagery-based smoke detection is at two levels, pixel and scene. Pixel level smoke detection, called smoke segmentation, gives a prediction for every pixel of an image to indicate whether it is a smoke pixel or not \citep{seg-threshold-99, labeldetectionCloud_6cls2016Hollstein, smoke-flumes-2021a, smoke-fire-segm-CNN-landcam-21, smoke-Unet-pix-2022, VIIRs-Landsat-imgs2023}. 
In contrast, scene level smoke detection aims to predict whether the scene in an image contains smoke or  not \citep{ smokenet_2019, zhaoliang22, ForestFireSmoke-Zheng2023,smoke-classif-random-forest-23,clssify-smoke-pyramid-Tao2024IoT} without concerning specific pixels.  A special type of scene level detection is  object detection which applies the methods for general object detection like YOLO, Mask R-CNN \cite{He_MaskRCNNBased2022, object-detection-smoke-VFnet-2024, Dou_ImprovedYOLOv5sFire2023, UAV-YOLOv7-2023, UAV-YOLOv8-Say23,smoke-landcam-Jin23sensors,cross-contrast-swin-transformer-2023} to detect smoke objects. This type of methods predict whether smoke presents in an image at the scene level, but at the same time, predicts a bounding box to show the locations of smoke objects. In comparison of the two types of methods, pixel level smoke detection has the advantages that the detection results are more interpretable as the scales of the smoke are clearly shown in the detection. At the same time, smoke locations can be derived from the detected smoke pixels and the locations are important for further action. This paper focuses on pixel level fire smoke detection, i.e., smoke segmentation. 

Early segmentation methods for smoke segmentation use thresholds \citep{seg-threshold-99,Zhao_DustSmokeDetection2010}. They apply thresholds from experiences to band readings or derivation of band readings to determine if a smoke pixel is smoky. To overcome the difficulty of acquiring thresholds for complex scenarios, machine learning methods and deep learning methods have been widely investigated. Machine learning methods apply predictive models like SVM, neural network, and random forest to predict if pixels are smoky after properties are derived for regions of an image. 
The work in \citep{Zhao_DustSmokeDetection2010} uses thresholds on bands readings and their derivatives to detect smoke and clouds. 
In \citep{forestSmokeNNModis2015} a multi-threshold method is used for extracting training sets to train a neural network classifier for merging the smoke regions.
The work in \citep{fire-Detect-threshold_distribu-20} proposes an adaptive threshold method using 2D Otsu Method to overcome the omission issue caused by fixed thresholds. 
The work in \citep{SVMSmokeDetection2019} proposes a method to iteratively combine superpixels and then use SVP to make smoke predictions with the aim of  overcoming the over-segmentation issue.  
The work \citep{smoke-superpixel-entropy-cluster-23} uses the superpixel technique to find regions where pixels are similar, use information entropy theory to find the  truncation distance, and then cluster centres. SVM is applied to make predictions for regions.  
The work in \citep{SmokeDetectionHimawari8-Mo2021} applies multi-layer perceptron (MLP) to the Kalimantan Island dataset. 
The work in \citep{Smoke-subpixel-Xu2024} is on subpixel analysis for smoke using RF, SVM, and threshold-weighted fusion.  
The work in \citep{SatelliteImagerySmoke-Sun2023} uses the distance between a pixel to its background distribution to decide if it is a smoke pixel.   
The survey paper \citep{survey-smoke-detect-tech2022} published in 2022 summaries more previous methods.

Deep learning methods use CNN and many types of attention mechanisms to predict whether a pixel is smoky directly without the preprocessing step for dimension reduction. The main advantage of deep learning methods is that they easily meet the high dimensionality challenge of images. These methods also  achieve higher  performance compared to other previous methods.  
The work in \citep{smoke-fire-segm-CNN-landcam-21} proposes a CNN model to segment smokes in RGB images from land cameras. A fully convolution network model is proposed in \citep{smoke-flumes-2021a} to detect smokes from satellite images. 
The Smoke-UNet model in \citep{smoke-Unet-pix-2022} applies a four level UNet model to smoke detection from multispectral imagery and achieves better results.  
The work in \citep{Yuan_SmokeSemanticSegmentation2023} proposes a CNN model that involve two residual paths with different sizes of convolutions to capture features of different scales. 
The model of \citep{Yuan_LightweightNetworkSmoke2023} extracts features via three stages with different resolution, passes these features through spatial attention and channel attentions, and finally makes segmentation predictions.   The model of \citep{Wen_DenseMultiScale2023} uses four stages and the output of the last stage are further sampled to reduce the resolution of feature maps. Then, all these features go through attention mechanisms for the final segmentation prediction.    The work \citep{SmokeSegm-DiscrimFeat-backg-foreg-synthesis-Tao2023PR} is on images from surveillance cameras where the smoke image and the background image without smoke are known. The work proposes a model to restore the smoke and its background given a smoke image. Features for background and for smoke are derived from a combined encoding phase but with separate decoding phase and supervised by the known background image and the known smoke respectively. Attentions on resolution and channels are heavily used in the work.     

Smoke segmentation shares some similarity with the segmentation of types of objects in remote sensing like buildings and roads and waters \citep{CNN-landcover-2019,  Sun_MA-UNetBuildingsWaters2022}, dust \citep{density-aware-DAUnet-dust-segm-2024,RandomForest-dust-identify2021} and fires \citep{seman-seg-fires-RAUnet-23, SpatioTemporalSelfAttention-video-fire-2022,fire-GFUnet-2024}.  UNet, which is also used in this paper, is used in \citep{seman-seg-fires-RAUnet-23,density-aware-DAUnet-dust-segm-2024,Sun_MA-UNetBuildingsWaters2022, fire-GFUnet-2024} for its good performances. RAUNet \cite{seman-seg-fires-RAUnet-23} for fire detection integrates residual blocks and attention gates at the decoding phase at every level of UNet. DAUNet \cite{density-aware-DAUnet-dust-segm-2024} for dust detection uses channel-spatial attention heavily in both ends of every level of UNet. This model is similar to GFUNet \citep{fire-GFUnet-2024} for fire detection. MA-UNet \citep{Sun_MA-UNetBuildingsWaters2022} for building detection combines self-attention based modules with UNet. MA-UNet uses UNet and self-attention and is closer to our method. We will present a comparison of our method with the closely-related latest GFUNet and MA-UNet in addition to Smoke-UNet.    

Existing methods/models do not perform well in detecting smoke pixels from complex satellite images despite the amount of work described above. The reasons for low detection performance are that smoke in satellite images may be with various  lighting conditions, smoke density, unclear borderline of thin smokes, background aerosols like clouds, haze, mist, or other special earth surface covers like forests, grass, bare land, constructions and roads, beaches, waters etc. Figure~\ref{fig:relativeness} shows some of these cases in our data set. Our data set contains Landsat images of six channels: RGB, NIR (near infrared), SWIR1 and SWIR2 (shortwave infrared). Some of these images contain thin smokes (areas labelled by A in (a) and (d) of the figure), some clouds and their shadows in (b), (c) and (d), unknown ground cover (c), smoke of different colors and different density, and on different backgrounds. 

A further reason for low detection performance is that the role of thin smoke varies based on its relation to the foreground (detection target) or background. Specifically. When thin smoke appears over a clear background, it acts as the detection target (e.g, Figure~\ref{fig:relativeness}(a)). When thin smoke is accompanied by thick smoke, it is no longer the detection target (e.g., the thin smoke around A in Figure~\ref{fig:relativeness}(b)), but forms a background and the target shifts to the thick smoke. This property of thin smoke is called contextual property and the features derived from it is called contextual features.       

These reasons lead to a question: can the performance of smoke detection/segmentation be improved when smoke appear in complex images? Our paper proposes a method to solve the problem raised by the question.  

Our method is a deep learning UNet-based model. UNet \cite{unet-ronneberger2015} has been widely used in segmentation of objects from images and in smoke, fire, land cover segmentation \citep{density-aware-DAUnet-dust-segm-2024, Sun_MA-UNetBuildingsWaters2022, seman-seg-fires-RAUnet-23,fire-GFUnet-2024,smoke-Unet-pix-2022} because its good performance as reviewed previously. Our model consists of a virtual channel construction  module to expand feature channels and a self-attention based vision transformer within Unet. The virtual channel construction  module expands input images to have a larger number of channels so that specific channels align to spectral patterns \cite{ZHAO2024-spectral-patterns}. The vision transformer is engineered to contrast the differences of region properties of region's means and maxima as done in \citep{SmokeSegm-DiscrimFeat-backg-foreg-synthesis-Tao2023PR}. Its self-attention extracts correlations of smoke relativity between regions and derive contextual correlation of thin smoke to make the model effective.     


The contributions of this paper consist of the following.
\begin{itemize}
\item  It proposes a machine learning model, called  VTrUNet that expands feature channels using a virtual channel construction  module to capture spectral patterns and captures long range and contextual feature associations using a transformer-boosted UNet. 

\item  It proposes a moderated F1 score for model evaluation and the score considers the quality of pixel labelling in complex condition and quality refers to the proportion of labelled pixels in contrast with the proportion not labelled.  
\item It presents a comprehensive ablation study of the different ways combining attention modules with UNet through an architecture framework to validate the performance of our model.  
\item It presents a comparison of our proposed model with the latest methods for and relating to smoke detection, such as GFUNet \cite{fire-GFUnet-2024}, MA-UNet \cite{Sun_MA-UNetBuildingsWaters2022}, Smoke-UNet \cite{smoke-Unet-pix-2022} and the CNN model in \citep{smoke-flumes-2021a}. The results show that our model is the best performer and achieves a clear improvement by more than 4\% from the second best model in terms of F1 score. 
\end{itemize}

\section{Method}

\newcommand\jxoC{\ensuremath{\bigcirc\hspace*{-1.7ex}c\hspace*{0.75ex} }}
\newcommand\jxoast{\ensuremath{\bigcirc\hspace*{-1.6ex}\ast\hspace*{0.5ex} }}
\newcommand\jxoplus{\ensuremath{\bigcirc\hspace*{-1.9ex}+\hspace*{0.4ex}} }
\newcommand\jxox{\ensuremath{\bigcirc\hspace*{-1.9ex}\times\hspace*{0.4ex} }}

\begin{figure*}[h]  \begin{center}
\includegraphics[scale=0.9]{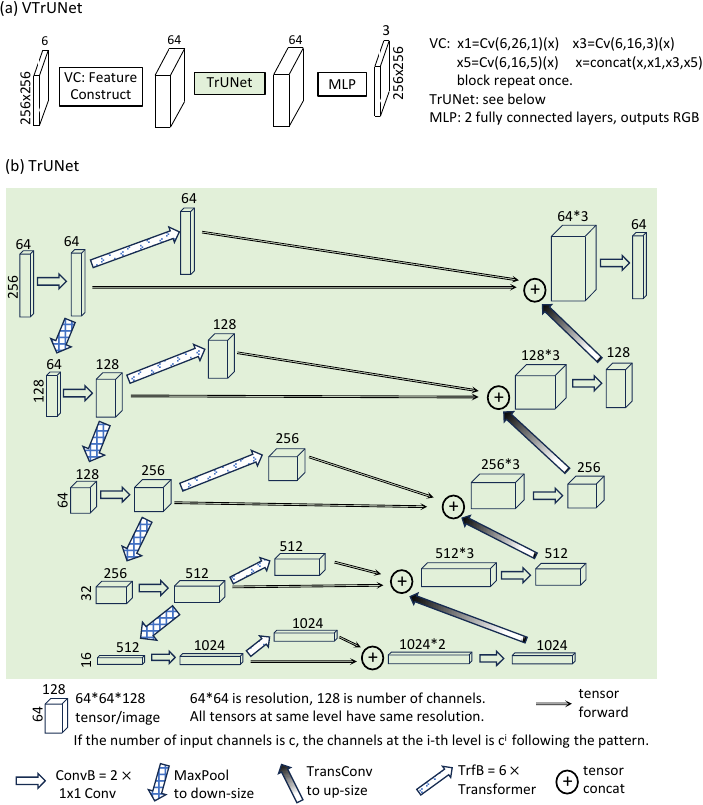}
\caption{\label{fig:arch}   (a) The architecture of the proposed model VTrUNet. (b) The transformer modified TrUNet where the transformer block TrfB is a stacking of vision transformer and will be described in the text. $Cv(c,u,s)$ is the convolution with c input channels, u output channels, and a kernel size of  s$\times$s.  TransConv means the transposed convolution to up-sample to increase the resolution of a feature tensor.}
\end{center}	
\end{figure*}

In this section, we first present the structure of our proposed model. Then, we show a metric for evaluating prediction performances of partially labelled images.  
 
\subsection{Our model VTrUNet}
Our method is a deep learning model that takes the input of images with 6 channels as described above and classifies each pixel in an image into three classes: Smoke, Cloud, and Clear. It is called VTrUNet, as shown in Figure~\ref{fig:arch}. It consists of two modules: a virtual channel construction module denoted by VC and a transformer-boosted UNet module denoted by TrUNet.  

The virtual channel construction (VC) module receives 6-channel input images and outputs tensors with 64 channels. Through channel expansion, the VC module represents different spectral patterns, such as objects in specific colors, in different channels for effective downstream learning \cite{ZHAO2024-spectral-patterns}. The output of the module is tensors with the same resolution as the input and consists of 64 channels. The selection of 64 channels is based on experiments.       

The VC module is also improved to use convolutions with kernel sizes 1x1, 3x3, and 5x5 to obtain features across different ranges. The features derived from these kernels are concatenated along the channel dimension. Simultaneously, the same set of kernels is repeated to capture features that might arise from non-linear interactions. In this design, features derived from areas larger than 5x5 are handled by the UNet and the transformer within the UNet. 

The transformer-boosted UNet, TrUNet shown in Figure~\ref{fig:arch}(b), is derived from the UNet model \cite{unet-ronneberger2015} by combining a  transformer block TrfB in every level of the UNet. At each level, the input image passes through the convolution block ConvB. The output image passes to the right hand side in two paths. In the top path, it goes through the transformer block TrfB to derive the long range relative features. In the bottom path at the level, the residual path goes directly to the right hand side. On the right hand side, the output of the transformer block, the output from the residual path, and the output of TransConv from the lower level are concatenated and go through the right ConvB to produce the output of the layer. The output of the left ConvB also goes downward to the next level through MaxPool2d down-sampling to reduce the resolution by half and to double the number of channels. On the right hand side, the output of the level goes upward through the transposed convolution (TransConv) to increase resolution and reduce the number of channels. The input to TrUNet is 256*256*c and the output is 256*256*c. If TrUNet follows the VC block, c is 64. If TrUNet is given the direct input images, c is 6. 

The transformer block TrfB is derived from the vision transformer (ViT) proposed in  \cite{ViT-ICLR2021Dosov} for classification in image data. It has been used in remote sensing for predictions \cite{SelfAttentionStrawSmoke2022,ForestFireSmoke-Zheng2023,forestfire-segm-FireViTNet-2024,Sun_MA-UNetBuildingsWaters2022}. In our implementation of the ViT, we use channel means and maxima (as in \citep{SmokeSegm-DiscrimFeat-backg-foreg-synthesis-Tao2023PR}) of regions as features for the self-attention in the ViT to derive the relevance of thin smoke to its surrounding regions. Specifically, the features help derive the contextual relationship of thin smoke in two contrasting settings. In the first setting, thin smoke is the detection target against a clear background. In the second setting, thin smoke serves as the background surrounding a thick smoke target.
Self-attention \cite{attentionIsAll2017} has been proven to one of the most successful techniques in capturing long-range associations. Different from previous work, we also use a residual connection parallel to the TrfB to the right hand side at each level. This residual connection enables the output of the left ConvB combined with the output of the TrfB in an optimal way at the right hand side. The transformer is repeated six times based on experiments, one after another, to capture super associations that require higher-order derivation.      

The MLP (multilayer perceptron) module for pixel prediction aims to predict the classes of the pixels in the image. It consists of two fully connected layers. The input to MLP is 256*256*c and the output is 256*256*3, an RGB image. The red channel corresponds to smoke, the green channel to clouds, and the blue channel to clear ground. This block can be easily adapted to problems where the number of classes to be predicted is more than three.  

We note that although UNet has been widely used in remote sensing-related segmentation \cite{smoke-Unet-pix-2022,fire-GFUnet-2024,Sun_MA-UNetBuildingsWaters2022,Filatov_ForestWaterBodies2022}, the closest model to our VTrUNet is MA-UNet in construction detection as both use a transformer at each level of UNet. Our model is different from MA-UNet in two ways. One is that we have a VC module to construct features and increase channels and the module contributes a lot to the performance as shown in the ablation study. Secondly, MA-UNet directs the output of the transformer at each level to the right hand side without a residual path while our model sends both the tensors of TrfB and the residual path to the right hand side. Our model allows the short-range features in the residual path and the long-range features in the output to be merged via the optimizable parameters in the network on the right side.     

We also note that spatial attention and channel attention are often used in deep learning models to escalate performances \cite{AttentionSmokeReview2021}. Spatial attention uses resolution compression and expansion to derive attention factors for the same pixels across all channels. In the case of UNet, this seems redundant as UNet has already had this compression on the encoding phase (left side) and expansion on the decoding phase (right side). Channel attention squeezes resolutions to derive attention factors for all pixels at specific channels.  In a later experiment from the ablation study presented later, we attached a channel attention module after TrUNet to improve the performance, but the performance gain from the channel attention was negative. More generally, our experiments show that the modules in an architecture are able to compensate for the role of a specific module when it is absent, and a module contributes to the performance depending on what other modules are in the model. More details will be presented in the ablation study section.

\begin{figure*}
\begin{center}
\includegraphics[scale=0.376]{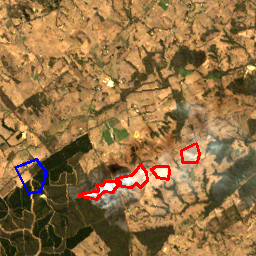}
\includegraphics[scale=0.376]{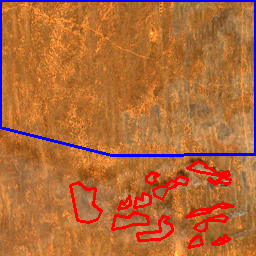}
\includegraphics[scale=0.376]{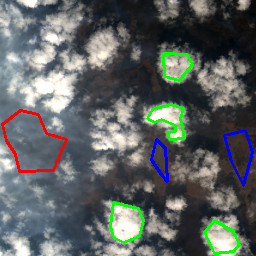}
\includegraphics[scale=0.376]{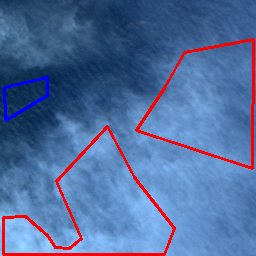}
\\ \vspace{0.5ex}
\includegraphics[scale=0.5]{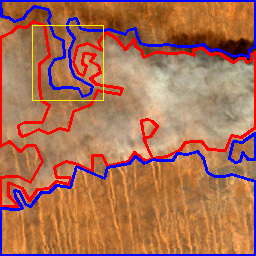}
\includegraphics[scale=0.5]{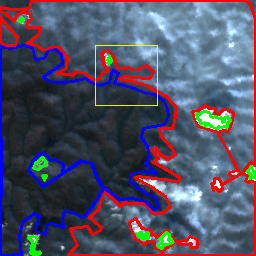}
\includegraphics[scale=0.5]{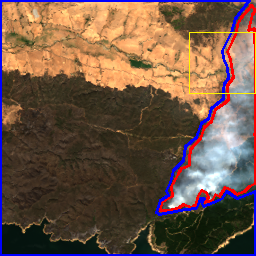}
\includegraphics[scale=0.5]{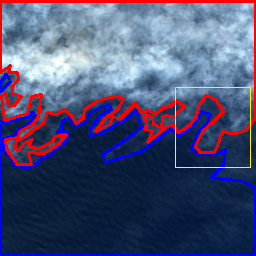}
\\
\includegraphics[width=3.4cm,height=3.4cm]{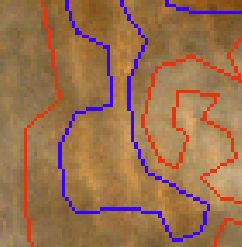}
\includegraphics[width=3.4cm,height=3.4cm]{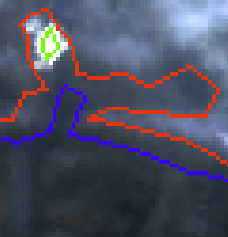}
\includegraphics[width=3.4cm,height=3.4cm]{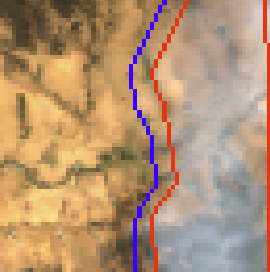}
\includegraphics[width=3.4cm,height=3.4cm]{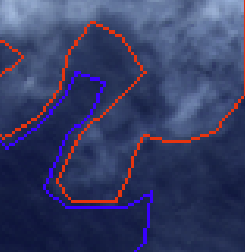}
\caption{Labelled images. Red polygons contain Smoke pixels, green Cloud pixels, and blue Clear pixels. Pixels between lines of different colors are unlabelled (gaps). The top row shows  training labels where typical pixels are labelled for training, leaving large areas unlabelled. The middle row shows the labelled images for evaluation where pixels are maximally labelled to reduce the unlabelled gap. The bottom row shows zoomed-in views of the respective middle row images, indicating that differentiating smoke pixels from non-smoke pixels is challenging in thin smoke.  
\label{fig:labelled-img} }
\end{center}
\end{figure*}

\subsection{A moderated F1 score for model evaluation}

Model training and evaluation requires labelled data. This requires borderlines to be drawn to indicate where smokes are in an image. Smoke may not have easily identifiable boundaries from its surrounding aerosols in a satellite image, as shown in the bottom row of Figure~\ref{fig:labelled-img}. This is typical in cases of thin smoke, smoke in low light conditions, and smoke in brownish backgrounds. In these cases, drawing borderlines to separate smokes from surrounding aerosols is almost an impractical task. A practical way to label an image is to label only the smoke, the clouds, and the clear backgrounds that the labeller is highly confident of and to leave the uncertain areas unlabelled. The uncertain areas are called gaps. This practical way of labelling is called partial labelling \citep{labeldetectionCloud_6cls2016Hollstein} \citep{AndersBook_RemoteSensing2021}(P.47). Figure  \ref{fig:labelled-img} shows a few images labelled for training (in the top row) and for evaluation (in the middle and the bottom rows).  To label images for training, only confident class pixels are labelled, but to label evaluation images, the images should be maximally labelled so that the model predictions in the unlabelled areas can be examined. We now present a moderation for evaluating prediction performances when gaps are present. 

We note that the metric Jaccard Index (Intersection Over Union - IoU) (as used in \cite{smoke-landcam-Jin23sensors}) is not capable to properly capture the prediction in the gap areas. In the spirit of IoU, if pixels labelled as Class A are predicted as Class A, they are true positives. Otherwise, they are errors. However, with partial labelling, when pixels in gaps are predicted as Class A, they are not errors. Only if pixels labelled as Class B are predicted as Class A are they errors. One might think to ignore the gap pixels to enable IoU calculation. Unfortunately, this can lead to a falsified performance. To make a model perform better, the labeller may label only a small number of easy-to-predict pixels for evaluation and not label any hard-to-predict pixels.

Our proposed modification considers gap sizes in addition to the evaluation for errors in the predictions. More specifically, we identify two conditions: (1) Predictions in the gap should not be counted as errors, and (2) smoke and/or cloud predictions in the labelled clear areas must still be counted as errors. We note that to differentiate the predictions in these two cases, clear pixels must also be labelled and predicted, like the smoke and cloud ones. 

\begin{figure}[h]
\begin{center}
\includegraphics[scale=1.]{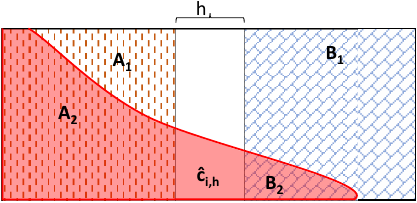}
\caption{\label{fig:score-func}
   The diagram for the metric. $A$ and $B$ are classes. $A_1$ is an area labelled as $A$. $\tilde{c}_i= A_1 \cup A_2$  are the labelled pixels of class $c_i$. $\tilde{c}_j=B_1\cup B_2$ are the labelled pixels of another class $c_j$. The middle vertical strip $h$ is the hazy gap and  contains all the unlabelled pixels. The red area consists of all the predicted $c_i$ pixels, $\hat{c}_i$. The red unshaded area $\hat{c}_{i,h}$ are the predicted $c_i$ pixels in $h$.
    }
\end{center}	
\end{figure}

We start from the standard performance metric of the standard F1 score to derive a moderated F1 score denoted by F1h to reflect the labelling quality. The definition of the standard F1 is given in Equations (1) where  $\tilde{c}_i$ denote all labelled pixels of class $c_i$, $h$ the gap (the unlabelled pixels), $\hat{c}_i$ all predicted pixels of class $c_i$, $\hat{c}_{i,h}$ the predicted $c_i$ pixels in the gap $h$, and $pn(x)$ the number of class $x$ pixels and $N$ all pixels of an image. The notation is illustrated in Figure~\ref{fig:score-func}. 
 \begin{align}
& prec(c_i) =  \frac{pn(\hat{c}_i \cap \tilde{c}_i )}{pn(\hat{c}_i) }  \qquad   &  
 rec(c_i) =  \frac{pn(\hat{c}_i \cap \tilde{c}_i)}{pn(\tilde{c}_i)  } \notag \\
 & F1(c_i) = \frac{2\cdot prec(c_i)\cdot rec(c_i)}{prec(c_i)+rec(c_i)} &
    \label{eq:pr-re-f1}
\end{align}


We design a modifier $r_h(c_i)=\frac{pn(\hat{c}_{i,h})}{pn(\hat{c}_i)}{+} \frac{pn(h)}{N}$ to reflect how well each class is labelled (in the first term) and how well the whole image is labelled (in the second term). When a class is well labelled, the first term is small as pixels of such a class would not be in the $h$ area. When most pixels in an image are labelled, the second term is small. Overall, a small modifier value $r_h(c_i)$ indicates better labelling quality.  

The moderated F1 score is Equation (\ref{eq:f1h}) below.

\begin{align}
F1h(c_i) =  F1(c_i)* (1-r_h(c_i))
    \label{eq:f1h}
\end{align}

It is well known that performance metrics should be evaluated for all classes. Otherwise, the metric values may not correctly reflect prediction quality. For example, if a dataset consists of two classes $P$ and $N$, where 10\% of the instances are class $P$ (positive) and the other 90\% are class $N$ (negative), and a model predicts all instances to be of class $N$, the recall for class $N$ by the model is 100\%, the precision for class $N$ is 90\%, and F1(N) = 0.95. These performance measurements are excellent. However, if class $P$ indicates the presence of a disease, the model fails to detect any disease cases. That is, the model fails to perform adequately. To overcome this problem, the performance metrics should be evaluated for all classes, and an average of the corresponding metrics should be used to represent model performance. Continuing the example, F1(P) = 0, and the average F1 score is 0.475, indicating a very poor model.

Following this approach, the metrics for all classes in an image are averaged to get the image-level metrics. The image-level metrics are then averaged over all images. The final averages of F1h, F1, precision, and recall across all classes and all images are the values we use to measure the performance of models on the dataset.

\section{Experiments and Results}
This section reports our experiment settings and results. 

\subsection{Data and labelling}
The dataset used in experiments is collected from Landsat 5 and Landsat 8 via the DEA Sandbox at https://docs.dea.ga.gov.au/setup/Sandbox/sandbox.html. The dataset contains 900 images for each of the classes: Smoke, Cloud, and Clear. Each image has a resolution of 256x256 and contains six channels: Blue, Green, Red, NIR, SWIR1, and SWIR2 with wavelength ranges of 0.63-0.69, 0.52-0.60, 0.45-0.52, 0.76-0.90, 1.55-1.75, and 2.08-2.35 respectively. Each pixel in an image represents 30 meters on the ground. More details about this dataset can be found in \citep{zhaoliang22}. The dataset is available on request.

Some images in the dataset, like Figure \ref{fig:pred-show}(i) and Figure \ref{fig:pred-dobetter}(iii), are not suitable for labelling because it is nearly impossible to know between smoke and cloud although the images are from a smoke area. From the dataset, 179 images are chosen for training and 14 images for evaluation. The images are chosen to reflect all types of pixels: heavy and thin smoke, white and black smoke, smoke above the sea, clouds, cloud shadows, in haze/mist, clear waters, and clear land covers of various kinds. Three classes are used in labelling: Smoke, Cloud, and Clear, with Clear meaning that the pixels are not covered by clouds or smoke. The training images are labelled for typical class pixels and have large areas of unlabelled regions as shown in the top row of Fig. \ref{fig:labelled-img}. The evaluation images are labelled to minimize the label gap as shown in the middle row of Fig. \ref{fig:labelled-img}. The numbers of pixels labelled for evaluation for the classes are respectively 100, 2, and 54 million. The labelled pixels for training are more than 10 times the number of  pixels for evaluation. The training images are augmented by rotations and flips.  

The labelling is done with Labelme \cite{labelme-wada2018}, a tool allowing polygons of different classes to be drawn on a image. The output is a JSON file describing the polygons. The polygons in the JSON file are then converted to patches of Smoke in red, Clouds in green, and Clear in blue in a PNG file. The PNG files are used as targets for training and evaluation.

\subsection{Model training} 

We trained many models as shown in the ablation study in the next section. The models are implemented in PyTorch. The loss function used in model training is the mean squared error of the predictions to the labelled pixels. The Adam optimizer was chosen to train the models. Dynamic learning rates are used with an initial value of 0.0001, which was decided by experiments. When training does not improve in 10 epochs, the learning rate is halved if it is not smaller than 1e-7. The training finishes if no improvement can be achieved in 20 epochs. Each model is trained with random initialization of parameters and is trained 10 times to find the best model. The training sessions of all the models achieved their best performance between 70 and 150 epochs. A GeForce GTX-1080 GPU with 10GB of memory is used in training. A training session of the VTrUNet model took about 2 hours. The batch size was set to 4, as our dataset and our model show a larger batch size slows down  model learning.

\subsection{Model performance and ablation study}

\begin{figure*}[h]  \begin{center}
\includegraphics[scale=0.9]{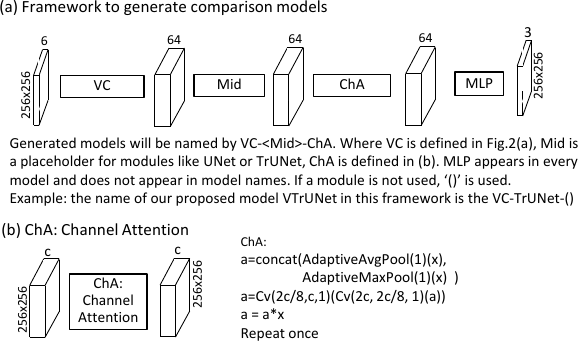}
\caption{\label{fig:mod-framework}   (a) The architecture of the model framework for ablation study. (b) channel attention module for context features. }
\end{center}	
\end{figure*}

Our ablation study follows the model framework in Figure~\ref{fig:mod-framework} to generate candidate models using the feature derivation module VC, the middle module Mid and the channel attention module ChA. The generated model will be named VC-$<$Mid$>$-ChA where $<$Mid$>$ is a place holder. An absent module is indicated by `()'. If all modules are absent from a model, the model's name is `()-()-()' which is the MLP model. In this case, the MLP model takes the 6-channel imagery on the far left side as input to predict the classes for pixels. If only VC is absent and $<$Mid$>$ is TrUNet, the model is ()-TrUNet-ChA, and  TrUNet takes the 6-channel imagery as input. The options for $<$Mid$>$ include TrUNet, UNet, UNet+TrfB. Here, UNet+TrfB means that $<$Mid$>$ is replaced by the sequence of the original UNet and the transformer block TrfB. This is different from TrUNet where TrfB is on an additional path at each level of the original UNet.   

\begin{table} 
\caption{models and their purpose for ablation study}\label{tbl:ablation-models}
\begin{center}
\begin{tabular}{ |l|l| }\hline
 model name & purpose   \\ \hline
 1:VC-TrUNet-()            &  this is our proposed model VTrUNet. \\  
 2:VC-TrUNet-ChA      &   to observe the effect of ChA \\
 3:()-TrUNet-()              &  to observe the effect of VC \\
 4:VC-()-()              &  to observe the effect of TrUNet \\
 5:VC-()-ChA        & to observe the effect of TrUNet with ChA \\
 6:VC-UNet-()              &  to observe the effect of TrfB \\
 7:VC-UNet+TrfB-() &  to observe the effect of TrfB \\
                                 &   when it is after UNet, instead of inside \\
  8:VC-UNet+TrfB-ChA & observe the effect of ChA \\
 9:()-()-() = MLP     &  to set a very basic model \\\hline
\end{tabular}
\end{center}
\end{table}

Based on the model framework, the models in our ablation study are listed in Table \ref{tbl:ablation-models}. The results of the experiments on models over the test dataset are listed in Table \ref{tbl:ablation-perform}. For each model, we report average F1h score (avgF1h) among all training sessions, and (F1h, F1, Precision,  recall) for the best model from these sessions in terms of F1h. The avgF1h score indicates the average performance of the model over all 10 training sessions. Model training involves random factors such random initial weights and random batches.    We also report the metric values for the best model because we always tend to choose to use the best-performing model if we have a choice. Models often balance the precision and recall differently, and F1 score indicates the results of the balance. The ratio F1h/F1 is an indicator of labelling quality. Better labelling leads to a high ratio value. 

\begin{table} 
\caption{models and their performances on the test data set}\label{tbl:ablation-perform}
\begin{center}
\begin{tabular}{ |l|c || c|c|c|c| }\hline
  &  & \multicolumn{4}{c|}{best}   \\ 
model id:name      &  avgF1h     & F1h & F1 & Prec & Rec    \\ \hline
1:VC-TrUNet-()      & {\bf 0.697}         &   {\bf 0.710} & {\bf 0.826} & {\bf 0.833} & 0.889      \\  
2:VC-TrUNet-ChA      & 0.684      &   0.693 & 0.812 &	0.792   & 0.901  \\
3:()-TrUNet-()              & 0.641      &   0.682 & 0.798 &	0.809 & 0.854      \\
4:VC-()-()                      &	0.625     &   0.626 &	0.712 &	0.771 &	0.766     \\
5:VC-()-ChA                &	0.652     &   0.7 &	0.808 &	0.832 &	0.832     \\
6:VC-UNet-()               &	0.657     &  0.667 &	0.786 & 0.754 &	0.872    \\
7:VC-UNet+TrfB-()     &	0.557    &  0.583 & 0.690 & 0.676 &	0.757        \\
8:VC-UNet+TrfB-ChA & 	0.549   & 0.625 & 0.728 & 0.725 & 0.796  \\
9:()-()-() = MLP             & 0.281   &  0.334 & 0.551 & 0.665 & 0.616           \\\hline
\end{tabular}
\end{center}
\end{table}

From the above comparison, we observe the following points. 
\begin{itemize}
\item{\bf The model VC-TrUNet-() performed best.} The F1h score, the F1 score and the precision are highest among all. The precision and the recall score are better balanced in contrast to `2:'. The average F1h score is also highest which makes this conclusion solid.    

\item{\bf The virtual channel construction model VC was very important.} By comparing 9:MLP and 4:VC-()-(), VC improves the performance a lot. This is also true when 1:VC-TrUNet-() and 3:()-TrUNet-() are compared. The comparisons show that virtual channel expansion is a very effective module to improve the performance of the model.  

\item{\bf The transformer boosted UNet, TrUNet, was very important.} This point is observed when 1:VC-TrUNet-() and 4:VC-()-() are compared.  The comparison shows that the transformer-boosted UNet is a very effective in improving the performance of the model.

\item{\bf Adding the transformer block TrfB to each level of UNet was more beneficial than adding it after UNet.} This point is observed when `1:VC-TrUNet-()' and `7:VC-UNet+TrfB-()' are compared and when `2:VC-TrUNet-ChA' and `8:VC-UNet+TrfB-ChA' are compared.  These comparisons demonstrate that the transformer-boosted UNet is the right choice compared to the UNet followed by the transformer. This is additional confirmation of the previous point about the transformer-boosted UNet. 

\item{\bf The contribution of a module to the performance of the model depends on what other modules are in the model.} If we compare `4:VC-()-()' with `5:VC-()-ChA', and compare `7:VC-UNet+TrfB-()' with '8:VC-UNet+TrfB-ChA', ChA's contribution is obvious. But when `1:VC-TrUNet-()' and `2:VC-TrUNet-ChA' are compared, ChA's contribution is negative. Same is true for VC. When  `1:VC-TrUNet-()' is compared with `3:()-TrUNet-()', the difference of F1h is 0.056.  When `4:VC-()-()' is compared with `9:()-()-()', the F1h difference is 0.344, much larger. This is an interesting result. It suggests  that \textbf{ adding more modules may not necessarily improve performance; in fact, it can worsen it. Additionally, the contribution of a module to the model performance depends on the other modules present in the model}. We just need to add the right ones to a model although which ones are right ones needs to be further investigated in the future. 
    
\end{itemize}

\subsection{Comparison to latest segmentation models}

\begin{table} 
\caption{A comparison of the performances of  VTrUNet  with the latest models.}\label{tbl:mod-comp}
\begin{center}
\begin{tabular}{ |l|c || c|c|c|c| }\hline
  &  & \multicolumn{4}{c|}{best}   \\ 
model      &  avgF1h     & F1h      &   F1     & Prec    & Rec    \\ \hline
SUNet      & 0.48            &  0.654  & 0.768  & 0.739 & 0.830      \\  
FrizCNN      & 0.541      &  0.585  & 0.691 &	0.666   & 0.77  \\
MA-UNet     & 0.586      &  0.626  & 0.725 &	0.713  & 0.835      \\
GFUNet       & 0.651       &  0.678  &	0.814 &	0.788 &	{\bf  0.889}     \\
VTrUNet     &{\bf 0.697}    &   {\bf 0.710} & {\bf 0.826} & {\bf 0.833} & {\bf 0.889}     \\\hline
\end{tabular}
\end{center}
\end{table}

In this subsection, we compare our model  VTrUNet  with recently proposed deep-learning segmentation models. The smoke segmentation  models are Smoke-UNet  \citep{smoke-Unet-pix-2022} (SUNet for short), FCN \citep{smoke-flumes-2021a}, and Frizzi's CNN (FrizCNN) \citep{smoke-flumes-2021a}. The architectures of SUNet is UNet based and FCN's architecture is very similar to UNet. Compared to FCN, SUNet has extra layers for channels 16 and 256. We choose to implement the newer and the deeper model, SUNet. We get the SUNet model code from \url{https://github.com/rekon/Smoke-semantic-segmentation}. We implemented the FrizCNN model ourselves as the paper supplies sufficient information for doing so. 

Recently, GFUNet \cite{fire-GFUnet-2024} was proposed for fire detection and MA-UNet \cite{Sun_MA-UNetBuildingsWaters2022} was proposed for buildings, roads and waters detection. Although they are not directly for smoke detection, we still compare them with our model to see how they perform on our data. We implemented the models based on the description in the papers.    

The input layers of these models were adapted to use 6 channels, and their output was adapted to 3 to align to our labelled data. Then, the training environment of the adapted models is the same as that of VTrUNet. The training parameters and the procedure are set the same as the those of training the  VTrUNet  model. Nevertheless, we tested and used best learning rate for these models to ensure that they are not at a disadvantage. 10 training sessions were run to get 10 trained models for each of them with our data.   

The performance values of the methods are shown in Table \ref{tbl:mod-comp}. The values show that  VTrUNet  achieved the highest performance, followed by GFUNet in second place, and MA-UNet in third. However, upon checking the predicted images, it was observed that thin smokes were often missed by these models. Interestingly, although SUNet's avgF1 score is low, its F1 scores across all training sessions vary dramatically. It achieved an F1 score of 0.654 in one training session, but dropped to a very low 0.401 in a few others.

\subsection{Effectiveness of  VTrUNet }

\begin{figure*}
\begin{center}
(i)\hspace{2.3cm}(ii)\hspace{2.3cm}(iii)\hspace{2.3cm}(iv)\hspace{2.3cm}(v)\\
\includegraphics[scale=0.30]{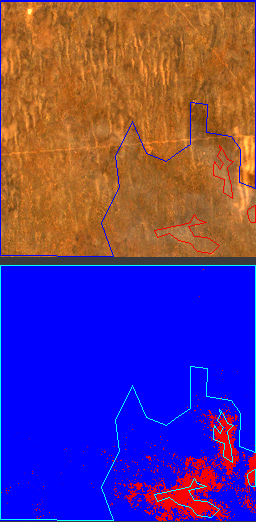}
\includegraphics[scale=0.30]{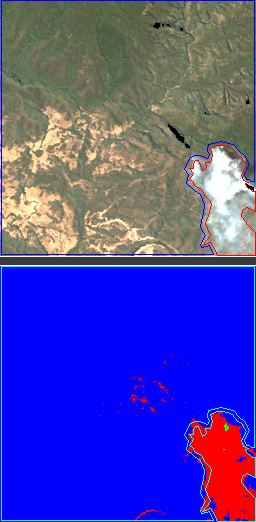}
\includegraphics[scale=0.30]{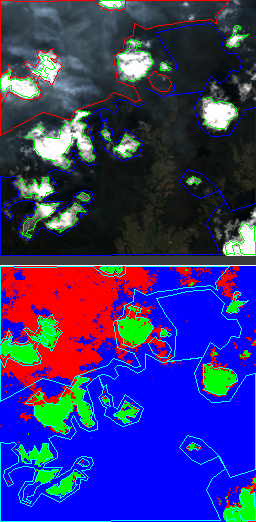}
\includegraphics[scale=0.40]{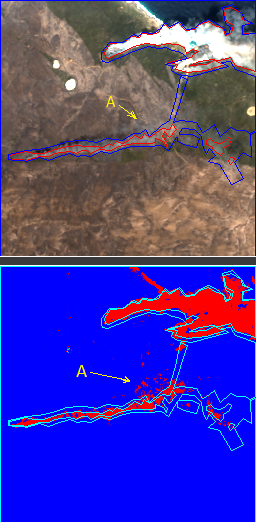}
\includegraphics[scale=0.30]{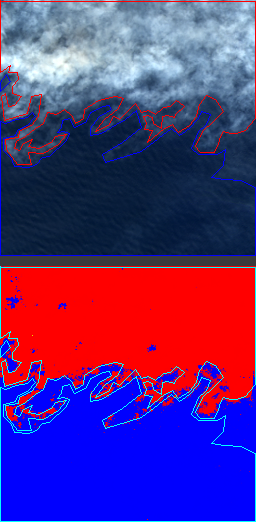}
\caption{\label{fig:test-show}
Test images and their predictions. Top row: the labelled images. Bottom row: pixel predictions by our model, red=Smoke, green=Cloud, and blue=Clear. In (ii), some brownish area is predicted smoke. This will be discussed more in the next section. (iv) Misses of thin smoke in area A in labelling. }
\end{center}
\end{figure*}

\begin{table} 
\caption{The evaluation scores for the images in Fig. \ref{fig:test-show}}\label{tbl:test-score}
\begin{center}
\begin{tabular}{ |l| c|c|c|c||c| }\hline
 image      &   F1h      &   F1     & Prec    & Rec    & F1h/F1\\ \hline
(i)         &  0.421	& 0.699	& 0.597	& 0.894  & 0.6    \\  
(ii)        &  0.88	& 0.923	& 0.948	& 0.903  & 0.95  \\
(iii)       &  0.664	& 0.858	& 0.831	& 0.896   & 0.77   \\
(iv)       & 0.819	& 0.875	& 0.934	& 0.835  & 0.94   \\
(v)        &   0.818	 & 0.907	& 0.922	& 0.902   & 0.90  \\\hline
\multicolumn{6} {r} {F1h/F1 indicates labelling quality}
\end{tabular}
\end{center}
\end{table}

\begin{figure*}
\begin{center}
(i)\hspace{2.3cm}(ii)\hspace{2.3cm}(iii)\hspace{2.3cm}(iv)\hspace{2.3cm}(v)\\
\includegraphics[scale=0.3]{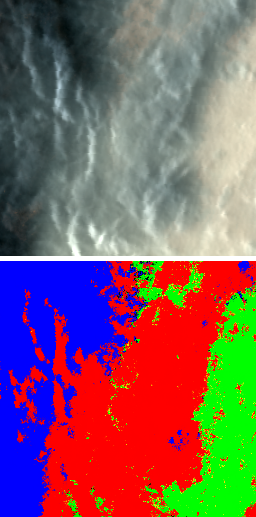}
\includegraphics[scale=0.3]{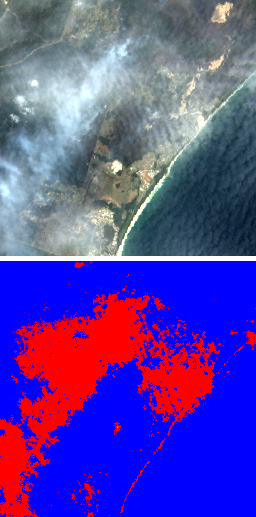}
\includegraphics[scale=0.3]{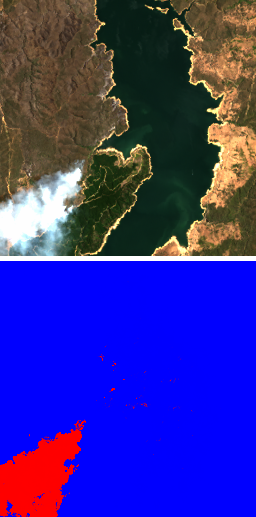}
\includegraphics[scale=0.3]{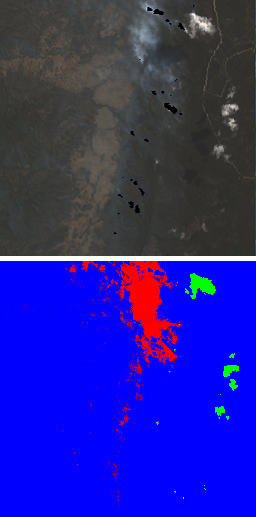}
\includegraphics[scale=0.3]{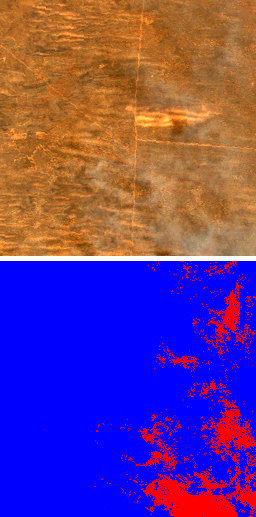}
\caption{\label{fig:pred-show}
Typical prediction examples of unlabelled images by VTrUNet. Top row shows RGB of the input images, and the bottom row shows the corresponding predictions. }
\end{center}
\end{figure*}

We visualize the performances of our model using examples  of test images labelled for evaluation  (Figure  \ref{fig:test-show}) and of unlabelled images (Figure \ref{fig:pred-show}). The example test images are shown in Figure \ref{fig:test-show} and their evaluation scores are shown in Table \ref{tbl:test-score}. Image (i) is well predicted with high recall, but the F1h score and the F1h/F1 ratio are low due to large unlabelled areas. Image (ii) achieves the highest F1h score. Although a small portion of the brown non-smoke area was mispredicted as smoke, the overall predictions are very accurate, and the labelling quality is also the best. Image (iii) presents a challenging case with many small patches of clouds, making border delineation difficult. However, our model performed well, accurately predicting all cloud patches and smoke areas. In Image (iv), the model correctly predicted thin smoke labelled with A, which was missed during labelling, in addition to correctly predicting other labelled pixels. Image (v) is also well predicted. 

From these evaluation images, the proposed model is effective. Smoke in different colors is detected correctly, and smoke and clouds are separated accurately. We believe that the accuracy and correctness are due to the spectral patterns captured by the virtual channel construction module and the long-range contextual features captured by the self-attention mechanism and the UNet.   

Figure \ref{fig:pred-show} displays examples of unlabelled images and the detected smoke in them by our model. Note that no evaluation score is available for these predictions because they are unlabelled. Case (i) illustrates smoke spreading to the valleys of a nearby region affected by bushfire, showcasing the model's ability to detect a combination of heavy smoke and thick clouds. Case (ii) demonstrates the detection of smoke with varying densities. Case (iii) presents an easy case showing accurate detection. Case (iv) highlights the model's capability to identify heavy smoke while disregarding thin smoke. Case (v) demonstrates the model's ability to identify thin smoke when contrasted with clear land.

The predictions for the unlabelled images validate that our model is powerful. It is able to distinguish smoke and clouds in complex conditions. It is able to correctly detect the role of thin smoke as a target and as background. We believe that this is due to contextual features and correlations captured by the self-attention in the UNet. 

Through visualization, we conclude that the overall performance of our model is highly accurate.

\subsection{Discussions}

\begin{figure}
\begin{center}
(i)\hspace{2cm}(ii)\hspace{1.7cm}(iii)
\\
\includegraphics[scale=0.35]{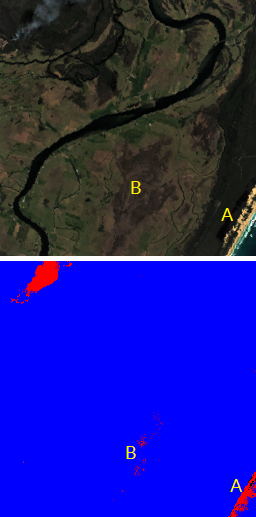}
\includegraphics[scale=0.35]{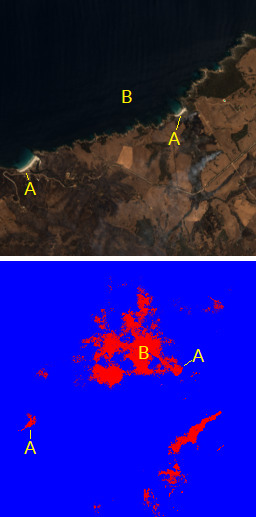}
\includegraphics[scale=0.2625]{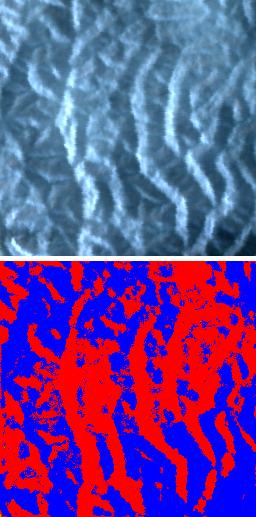}
\caption{\label{fig:pred-dobetter}
Predictions for discussion. Top row shows RGB of the input images, and the bottom row shows the corresponding predictions.  }
\end{center}
\end{figure}

Figure \ref{fig:pred-dobetter} presents images where the predictions are not perfect. The model's predictions in dark brown areas contain false positive errors. In Image (i), Area B is dark brown, resembling a burnt area, but the model incorrectly predicted it as smoke. Additionally, in the lower right corner of the image, the model misclassified beaches as smoke. Similarly, in Image (ii), beaches indicated by A were incorrectly predicted as smoke. In Image (iii), although smoke is spread in valleys, the detection only identified smoke on the sunny sides of the slopes.

We believe that these shortcomings can be addressed with more training data. However, the current dataset lacks images to adequately train the model for these special areas, especially for areas like burnt areas and shadowy slopes.

\section{Conclusion}
In this paper, we proposed a UNet and transformer based deep learning model to detect smoke under complex conditions. The model enables smoke to be detected based on features and contextual feature associations in long ranges. With this model, it becomes possible to detect different coloured smoke when it is light against a clear background, thick in a hazy background, and obscured by clouds and cloud shadows. We also proposed a metric for evaluating models when the evaluation images are labelled with the partial labelling method. This metric considers whether the labelling has labelled all typical pixels of an image to minimize the label gap. The experiments of this work show that the proposed method is very effective compared with the latest methods in the literature. 

This work needs improvement in a few areas. Firstly, work is needed to correct false positive predictions in sand beaches. Secondly, work is needed to correct false negative predictions in large area of shadows, and thirdly, work is needed to differentiate brown earth surfaces from burnt areas. These directions will be addressed in future work.  
 

\bibliographystyle{abbrv}
\bibliography{smoke-refs}

\end{document}